\documentclass[letterpaper]{article} 
\usepackage{aaai25}  
\usepackage{times}  
\usepackage{helvet}  
\usepackage{courier}  
\usepackage[hyphens]{url}  
\usepackage{graphicx} 
\usepackage[numbers]{natbib}  
\usepackage{caption} 
\usepackage{algorithm}
\usepackage{algorithmic}
\usepackage{amsmath}
\usepackage{amssymb}

\usepackage{newfloat}
\usepackage{listings}
\DeclareCaptionStyle{ruled}{labelfont=normalfont,labelsep=colon,strut=off} 
\floatstyle{ruled}
\newfloat{listing}{tb}{lst}{}
\floatname{listing}{Listing}

\usepackage{booktabs}
\usepackage{multirow}

\nocopyright
\title{Modeling Habitat Shifts: Integrating Convolutional Neural Networks and Tabular Data for Species Migration Prediction}
\author{
    Min-Hong Shih\equalcontrib,
    Emir Durakovic\equalcontrib
}
\affiliations{
    \textsuperscript{\rm}
    Northeastern University \\
    360 Huntington Ave \\
    Boston, MA 02115 \\
    durakovic.e [at] northeastern.edu \\
    shih.minho [at] northeastern.edu
}

\begin{document}

\maketitle

\begin{abstract}
Due to climate-induced changes, many habitats are experiencing range shifts away from their traditional geographic locations \cite{piguet2011}. We propose a solution to accurately model whether bird species are present in a specific habitat through the combination of Convolutional Neural Networks (CNNs) \cite{oshea2015} and tabular data. Our approach makes use of satellite imagery and environmental features (e.g., temperature, precipitation, elevation) to predict bird presence across various climates. The CNN model captures spatial characteristics of landscapes such as forestation, water bodies, and urbanization, whereas the tabular method uses ecological and geographic data. Both systems predict the distribution of birds with an average accuracy of 85\%, offering a scalable but reliable method to understand bird migration.
\end{abstract}
\section{Introduction}
Climate change has resulted in the relocation of various species \cite{mcdonald2011optimal}; for example:
\begin{itemize}
\item Bird species are shifting their migratory routes to higher longitudes to find suitable nesting areas \cite{lemoine2003potential}.
\item Invasive aquatic animals favor warmer waters where the success of cold water invasive species is minimized \cite{rahel2008assessing}.
\item Reptiles and amphibians are finding locations that match their thermal and moisture needs \cite{bickford2010impacts}.
\end{itemize}
We will tackle the first mentioned example above. Due to mobility, visibility, and well-documented historical ranges, birds have always been a good bio-indicator of environmental change \cite{morrison1986bird}. Hence, understanding how bird species respond to climate change is critical to conserving our forests and introducing future environmental policies on undoing global warming effects. 

Traditional methods of analyzing bird distributions usually rely on manual observations, a process that has major geographic coverage, temporal consistency, and resource limitations \cite{sullivan2017using}. A field biologist can only be in one location at a time. Hence, there is a growing need for more scalable approaches, such as modeling migration patterns through the use of predictive AI agents. A method that is often used to model species distributions is climate envelope models.

Climate envelope models are a subset of species distribution models that use climate variables such as climate, land cover, and topography to predict the environmental suitability of a species \cite{pearson2003predicting}. These models often use mathematical functions to describe the associations between variables and the presence of the species \cite{guisan2006making}. However, the limitations of climate envelope models lie in their inability to account for factors such as species interactions, dispersal limitations, and landscape features visible in satellite imagery \cite{ferrier2004mapping}. 

In this paper, we examine ways to classify whether birds will be present in various climate types. More specifically, we focus on implementing a CNN model that takes satellite imagery and predicts species migration patterns through local landscape features that would affect the abundance of a bird species, such as bodies of water, forestation, and urbanization. 

We also suggest a tabular model that generates a dataset that contains features such as latitude, longitude, elevation, precipitation, and temperature. This data set is then fed into a random forest classifier that determines whether a bird is present or absent in a given location. 

Our main contributions include: 
\begin{itemize} 
\item A CNN-based method for using satellite imagery to predict species presence across landscapes. 
\item A feature-driven random forest classifier that uses climate and topography data. 
\item A comparative analysis of both models using real-world bird occurrence data from the eBird database and pre-built models. \end{itemize}
Our research study offers a more scalable technique to predict bird distributions and hope to inspire better climate conservation practices.

\section{Background}
In this section, we will be discussing how the models used in this paper were constructed.
\begin{itemize}
\item \textbf{Random Forests} \\
A decision tree is represented by a node of a test attribute, and each corresponding branch is the result of that test. The random forest classifier constructs multiple decision trees and takes the average to decide the probability of a bird species being present at a given location \cite{breiman2001random}. 
\item \textbf{Gradient Boosting Decision Trees} \\
Similar to random forests, Gradient Boosting Decision Trees (GBDT) build trees sequentially with each new tree correcting the errors of the previous trees. We employ this technique as it generally performs better on smaller datasets due to its ability to reduce bias \cite{si2017gbdt}. 
\item \textbf{Convolutional Neural Networks} \\
CNNs are designed to extract spatial features from image data. In our case, satellite imagery is processed through a CNN to capture important landscape features such as water bodies, vegetation, and urban structures that may influence bird presence \cite{oshea2015}.
\item \textbf{Residual Network (ResNet)}
Residual Network, otherwise known as ResNet are a type of neural network architecture that bypasses layers. The "skip connections" allow for a more direct information flow across the network eliminating the issue of "vanishing gradients" that occur when training on a model with hundreds of layers. ResNets learn the "residual" hence their name, between the desired output and the layer input. This improves its performance for tasks such as object detection and image classification \cite{he2016resnet}. 
\item \textbf {Multi-label Classification}
Since multiple bird species can be present in the same region, we frame the problem as a multi-label classification task. The model outputs a vector of presence probabilities for each species, rather than a single-class label, enabling simultaneous prediction of multiple species.
\end{itemize}
\section{Related Work}

Similar research projects have explored both bird migration distributions and habitat changes over several years. We begin by exploring the SatBird research study that generates a dataset based off of Sentinel-2 \cite{esa2023sentinel2} satellite imagery, biological- and climate-oriented rasters, and the eBird dataset \cite{sullivan2009ebird} for presence labeling. This dataset is divided into USA-summer, USA-winter, and Kenya-specific regions. The research project then inputs their dataset into baseline model classification and presents the dataset for further scientific development \cite{teng2023satbird}.

We also looked into a research project dealing with habitat prediction for spatial distribution of Japanese rice fish. The researchers of this project used Fuzzy Neural Networks, a system which mimics human cognitive abilities with Neural Networks to handle uncertainty and learn from the data, among other models. Their dataset involves features like depth and vegetation coverage, data which is then fed into the prediction model to evaluate habitat preference \cite{fukuda2013fish}.

As discussed, there have been existing studies that have made efforts to species distribution modeling; however, limitations remain. The SatBird dataset, although comprehensive in its integration of satellite imagery with citizen science data, focuses primarily on static seasonal distributions and lacks the temporal resolution to capture progressive movement patterns. Studies using Fuzzy Neural Networks for habitat prediction show strong uncertainty that offer little transferability between areas. Our study addresses this gap \cite{teng2023satbird} \cite{fukuda2013fish}.

\section{Methods}

We define our problem as such: given a location (\( x_{lat} \), \( x_{lon} \)), where \( x_{lat} \) and \( x_{long} \) represent the latitude and longitude coordinates respectively, with its respective features, H, and a binary label \( y_i \) generate a dataset, D(H, \( y_i \)), such that we can learn a function f(D) that minimizes the classification error for presence prediction across a range of birds with output, z.

Currently, observations are simply recorded when a bird is spotted at a certain location. Manually transcribing observations takes effort and can grow tiring. We can further simplify manual classification by combining well-known datasets to create a more comprehensive dataset. We propose a solution to utilize this exhaustive dataset to generate predictions further enhancing bird distribution models. This exhaustive dataset generated through the longitude and latitude would not only contain tabular data similar to data generated through manual transcription, but would also contain satellite images of the longitude and latitude area. By fine-tuning our dataset to consider only necessary features that impact birds we can offer reliable and more effective classifications than traditional models. 

\subsection{Datasets}
\begin{itemize}
\item \textbf{eBird Dataset} \\
We use the Cornell Lab's eBird dataset \cite{sullivan2009ebird} to pull bird observation data. This dataset contains the latitude, longitude, observation date, and presence data of multiple bird species in Northern America. 

\item \textbf{WorldClim Dataset} \\
The WorldClim dataset \cite{fick2017worldclim} features raster files with locations spaced at a resolution of one square kilometer. Each coordinate maps to a real-world location and contains information like its latitude, longitude, elevation, precipitation, and temperature depending on the downloaded dataset .

\item \textbf{Sentinel-2 Dataset} \\
This dataset provides satellite imagery using Microsoft's Planetary Computer API. Using longitude and latitude, a satellite image of the area can be queried in a variety of bandwidth, from human visible wavelengths to non-visible ones. Through the use of queries to the API, it can be specified to a specific condition of weather like "no cloud condition" for a clear image of the landscape \cite{esa2023sentinel2}.

\end{itemize}
With regards to the tabular model, we combine the eBird and WorldClim datasets to generate a representative dataset of features like: latitude, longitude, elevation, precipitation, temperature, but also the bird species type and its presence at that location. Since eBird only provides presence observations, we generate pseudo-absent observations by checking whether a location is not too close to a known observed location for that bird (within a 1.1km radius).

\subsection{Statistical Models}
We divide our dataset into a 70:10:20 split for training, validation, and testing, respectively\cite{Cutler2007}.
\subsubsection{Tabular Model}
To model species occurrence, we use a random forest classifier trained on a tabular dataset combining environmental features from WorldClim and species observations from eBird. Since eBird only provides presence data, we generate pseudo-absence points from ecologically plausible but unobserved regions\cite{Barbet-Massin2012}. This allows the model to distinguish between suitable and unsuitable habitats.

The random forest builds multiple decision trees using bootstrap samples and feature bagging\cite{Prasad2006}. Each tree outputs a probability estimate for species presence, and the final prediction is the average of all trees. We apply a threshold $\theta$ (that is usually 0.5) on the validation set to convert probabilities into binary predictions, to evaluate our model for accuracy. We can then simply return the probabilistic estimate for the prediction of bird presence. The corresponding pseudo-code is presented in Algorithm~\ref{alg:train_rf}.

In addition to random forests, we experiment with gradient boosting (e.g., XGBoost), a model which trains trees sequentially on residual errors from previous models. Boosting should improve accuracy for smaller datasets, but for our case it may struggle to classify due to the noise in an ecological setting. Still, we use this model to experiment and contrast between the two. See Algorithm~\ref{alg:gbdt}.

We evaluate model performance using our 20\% testing data split. Moreover, we analyze feature importance to identify key environmental drivers of species distribution\cite{Mi2017}. The steps for evaluating this model are detailed in Algorithm~\ref{alg:predict_rf} and Algorithm~\ref{alg:eval_gbdt}.

We chose these two models to compare how a more generalized model contrasts with a sensitive to noise model. Especially since we are using smaller datasets, we want to evaluate how the two use the features to differentiate whether a bird is present or absent at a given location.

\subsubsection{Convolutional Neural Network}
For the ResNet CNN approach, we implemented a transfer learning approach using a modified ResNet-34 architecture\cite{He2015}. The model extracts features from Sentinel-2 RGB imagery (bands B02, B03, B04) at a resolution of 10m by using pre-trained weights from ImageNet. The backbone for this remained the ResNet-34 architecture, while the final fully connected layer was substituted with a custom classifier for multi-label species prediction. 

We implemented a dropout of (p = 0.5) in the classification head and froze early convolutional layers to reduce overfitting in our small ecological dataset. Binary cross-entropy loss with Adam optimization was used to train the model (weight decay = 1e-5, initial lr = 1e-4)\cite{Kingma2014}. Given the intrinsic variability in satellite imagery and the spatial correlation in ecological observations, these methods successfully strike a balance between fitting the training data and generalizing to unseen habitats when paired with weight decay. The pseudo-code is shown in Algorithm~\ref{alg:resnet_cnn}.

Next a custom CNN was implemented in comparison with the ResNet approach. Without using pre-trained weights, our unique CNN model was created especially for predicting bird habitat. Each of the four convolutional blocks that make up the architecture has two convolutional layers with batch normalization, ReLU activation, and max pooling. The filter depths of the blocks range from 64 to 128 to 256 to 512. Sentinel-2 satellite imagery features, which are very different from those of natural photography datasets like ImageNet, were the focus of this configuration's optimization. To guarantee appropriate gradient flow through the deep network, we used Kaiming initialization\cite{He2015a}. With a learning rate of 1e-4 and a weight decay of 1e-5 for regularization, the model was trained end-to-end using binary cross-entropy loss and Adam optimization. The pseudo-code is shown in Algorithm~\ref{alg:custom_cnn}.

\begin{algorithm}[tb]
\caption{Train Random Forest}
\label{alg:train_rf}
\textbf{Input}: Feature matrix $X$, labels $y$ \\
\textbf{Parameter}: Number of trees $T$, max depth $d$, min samples split $s$, max features $f$ \\
\textbf{Output}: List of trained trees
\begin{algorithmic}[1]
\STATE Initialize list $trees \leftarrow [\ ]$
\FOR{$i = 1$ to $T$}
    \STATE $(X_{sample}, y_{sample}) \leftarrow$ SampleWithReplacement($X, y$)
    \STATE $tree \leftarrow$ TrainDecisionTree($X_{sample}, y_{sample}, d, s, f$)
    \STATE Append $tree$ to $trees$
\ENDFOR
\STATE \textbf{return} $trees$
\end{algorithmic}
\end{algorithm}

\begin{algorithm}[tb]
\caption{Predict with Random Forest}
\label{alg:predict_rf}
\textbf{Input}: Test data $X_{test}$, trained trees $trees$ \\
\textbf{Parameter}: Threshold $\theta$ \\
\textbf{Output}: Predictions and average probabilities
\begin{algorithmic}[1]
\STATE Initialize list $all\_probs \leftarrow [\ ]$
\FOR{each $tree$ in $trees$}
    \STATE $probs \leftarrow$ PredictProba($tree, X_{test}$)
    \STATE Append $probs$ to $all\_probs$
\ENDFOR
\STATE $avg\_probs \leftarrow$ Average($all\_probs$, axis=0)
\STATE $predictions \leftarrow [1$ if $p \geq \theta$ else $0$ for $p$ in $avg\_probs]$
\STATE \textbf{return} $(predictions, avg\_probs)$
\end{algorithmic}
\end{algorithm}

\begin{algorithm}[tb]
\caption{Gradient Boosting Training}
\label{alg:gbdt}
\textbf{Input}: Feature matrix $X$, target labels $y$ \\
\textbf{Parameter}: Number of trees $T$, learning rate $\eta$ \\
\textbf{Output}: Final model $F_T(x)$
\begin{algorithmic}[1]
\STATE Initialize model prediction: $F_0(x) =$ mean of $y$
\FOR{$t = 1$ to $T$}
    \STATE Compute residuals: $r_i = y_i - F_{t-1}(x_i)$
    \STATE Train a regression tree $h_t(x)$ on inputs $X$ and residuals $r$
    \STATE Update model: $F_t(x) = F_{t-1}(x) + \eta \cdot h_t(x)$
\ENDFOR
\STATE \textbf{return} final model $F_T(x)$
\end{algorithmic}
\end{algorithm}

\begin{algorithm}[tb]
\caption{Evaluate Gradient Boosting}
\label{alg:eval_gbdt}
\textbf{Input}: Test data $X_{test}$, true labels $y_{true}$, trained model $F_T(x)$ \\
\textbf{Parameter}: Threshold $\theta$ \\
\textbf{Output}: Binary predictions and evaluation metrics
\begin{algorithmic}[1]
\STATE Compute predicted probabilities: $probs \leftarrow F_T(X_{test})$
\STATE Initialize empty list $predictions \leftarrow [\ ]$
\FOR{each $p$ in $probs$}
    \IF{$p \geq \theta$}
        \STATE Append 1 to $predictions$
    \ELSE
        \STATE Append 0 to $predictions$
    \ENDIF
\ENDFOR
\STATE Compare $predictions$ with $y_{true}$ to compute:
\STATE \quad Accuracy, Precision, Recall, F1-score, AUC
\STATE \textbf{return} Evaluation metrics
\end{algorithmic}
\end{algorithm}

\begin{algorithm}[tb]
\caption{ResNet-based Bird Habitat CNN}
\label{alg:resnet_cnn}
\textbf{Input}: Sentinel-2 RGB imagery $X \in \mathbb{R}^{3\times224\times224}$, species presence labels $y \in \{0,1\}^n$ \\
\textbf{Output}: Species presence probabilities $p \in [0,1]^n$
\begin{algorithmic}[1]
\STATE Load pre-trained ResNet-34 with ImageNet weights
\FOR{each parameter in ResNet layers[1:len(layers)-8]}
    \STATE parameter.requires\_grad $\leftarrow$ false  \COMMENT{Freeze early layers}
\ENDFOR
\STATE Replace ResNet.fc with Identity layer
\STATE Define classifier as Sequential(
\STATE \quad Dropout(0.5),
\STATE \quad Linear(512, 256),
\STATE \quad ReLU(),
\STATE \quad Dropout(0.5),
\STATE \quad Linear(256, $n$),
\STATE \quad Sigmoid()
\STATE )
\STATE \textbf{function} forward($X$)
    \STATE features $\leftarrow$ backbone($X$)  \COMMENT{Extract features using modified ResNet}
    \STATE $p \leftarrow$ classifier(features)  \COMMENT{Predict species probabilities}
    \STATE \textbf{return} $p$
\STATE \textbf{end function}
\end{algorithmic}
\end{algorithm}

\begin{algorithm}[tb]
\caption{Custom CNN for Bird Habitat Prediction}
\label{alg:custom_cnn}
\textbf{Input}: Satellite imagery $X \in \mathbb{R}^{c\times224\times224}$, labels $y \in \{0,1\}^n$ \\
\textbf{Output}: Species probabilities $p \in [0,1]^n$
\begin{algorithmic}[1]
\STATE features $\leftarrow$ Sequential(
\STATE \quad Conv2d($c$, 64, kernel\_size=7, stride=2, padding=3),
\STATE \quad BatchNorm2d(64), ReLU(), MaxPool2d(3, stride=2),
\STATE \quad Conv-BN-ReLU(64, 128) $\times$ 2, MaxPool2d(2),
\STATE \quad Conv-BN-ReLU(128, 256) $\times$ 2, MaxPool2d(2),
\STATE \quad Conv-BN-ReLU(256, 512) $\times$ 2, MaxPool2d(2)
\STATE )
\STATE 
\STATE classifier $\leftarrow$ Sequential(
\STATE \quad AdaptiveAvgPool2d((1, 1)), Flatten(),
\STATE \quad Dropout(0.5), Linear(512, 256), ReLU(),
\STATE \quad Dropout(0.5), Linear(256, $n$), Sigmoid()
\STATE )
\STATE 
\STATE \COMMENT{Apply Kaiming initialization to all layers}
\FOR{each module $m$}
    \STATE Apply appropriate initialization based on layer type
\ENDFOR
\STATE 
\STATE \textbf{function} forward($X$)
    \STATE $X \leftarrow$ features($X$)
    \STATE $p \leftarrow$ classifier($X$)
    \STATE \textbf{return} $p$
\STATE \textbf{end function}
\end{algorithmic}
\end{algorithm}

\section{Results}

\subsection{Tabular Data}
We evaluate our models based off accuracy which computes the proportion of all predictions that were correct and the "Area Under the Curve of the Receiver Operating Characteristic curve" (AUC-ROC), a measurement which shows how well a model separates data. We choose these two metrics because birds may be rarer in some regions than others, hence AUC helps us understand whether the model is learning relevant features from the dataset and accuracy shows how well our models are performing in general. 

Our Random Forest and Gradient Boosting models will be compared alongside the baseline Scikit-learn implementations of the two mentioned models. We compare both the validation and test accuracies to understand both the generalized performance and hyperparameter choices of our models and our actual accuracy of the model. For demonstration purposes, we will be training on four different bird species—American Robin, Pileated Woodpecker, Blue Jay, and Carolina Wren. For each species, we sample 250 presence and 250 pseudo-absence statistics respectively. The summarized report for each bird species and their respective models can be found in Table~\ref{tab:results}. As shown, our Random Forest (RF) model performs with approximately 86\% validation and lingers around 80\% testing accuracy respectively. We also conclude that our model is separating the two classes with a high accuracy of 92\% validation and 84\% testing AUC accuracy. Our Gradient Boosting Tree (GBT) model performs with a slightly worse accuracy, an observation which presents some interesting facts about our dataset. 

\begin{table}[t]
\centering
\small  
\caption{Validation and Test Accuracy/AUC for Each Species and Model}
\label{tab:results}
\begin{tabular}{@{}llcc@{}}
\toprule
\textbf{Species} & \textbf{Model} & \textbf{Val Acc / AUC} & \textbf{Test Acc / AUC} \\
\midrule
\multirow{4}{*}{\shortstack[l]{American\\Robin}}
& RF               & 0.857 / 0.924 & 0.803 / 0.842 \\
& GBT              & 0.771 / 0.848 & 0.732 / 0.722 \\
& SK-RF            & 0.857 / 0.938 & 0.789 / 0.809 \\
& SK-GBT           & 0.886 / 0.942 & 0.831 / 0.785 \\
\midrule
\multirow{4}{*}{\shortstack[l]{Pileated\\Woodpecker}}
& RF               & 0.829 / 0.872 & 0.873 / 0.895 \\
& GBT              & 0.857 / 0.768 & 0.873 / 0.926 \\
& SK-RF            & 0.829 / 0.910 & 0.916 / 0.914 \\
& SK-GBT           & 0.800 / 0.884 & 0.887 / 0.864 \\
\midrule
\multirow{4}{*}{Blue Jay}
& RF               & 0.914 / 0.958 & 0.845 / 0.837 \\
& GBT              & 0.771 / 0.720 & 0.747 / 0.801 \\
& SK-RF            & 0.914 / 0.962 & 0.817 / 0.869 \\
& SK-GBT           & 0.886 / 0.936 & 0.761 / 0.843 \\
\midrule
\multirow{4}{*}{Carolina Wren}
& RF               & 0.971 / 0.942 & 0.884 / 0.885 \\
& GBT              & 0.941 / 0.981 & 0.899 / 0.872 \\
& SK-RF            & 0.941 / 0.904 & 0.870 / 0.859 \\
& SK-GBT           & 0.882 / 0.942 & 0.884 / 0.883 \\
\bottomrule
\end{tabular}
\end{table}

\begin{figure}[H]
\centering
\includegraphics[width=1\linewidth]{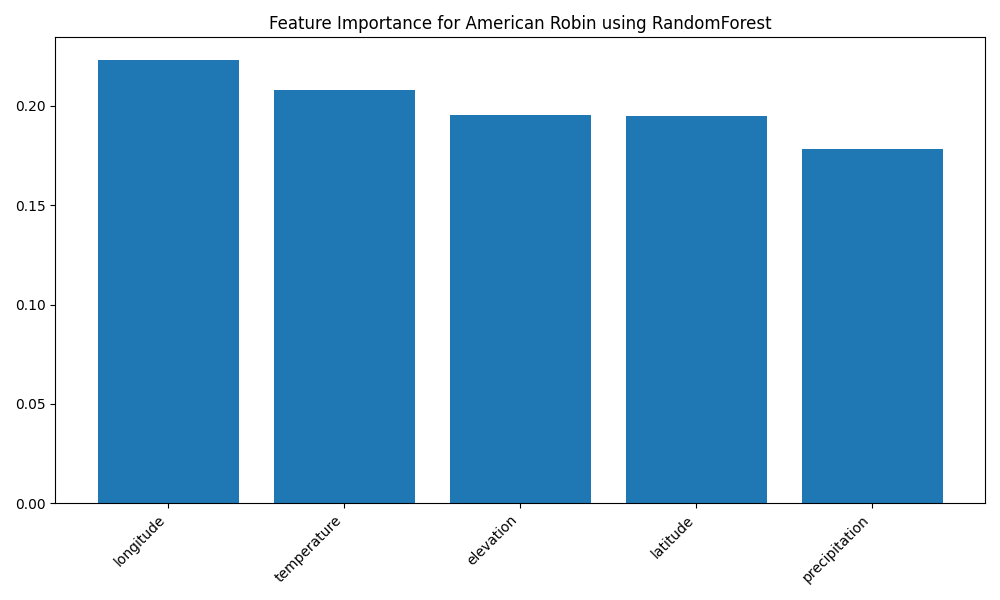}
\caption{Feature importance graph for the Random Forest model.}
\label{fig:robin_rf_feat_imp}
\end{figure}

\begin{figure}[H]
\centering
\includegraphics[width=1\linewidth]{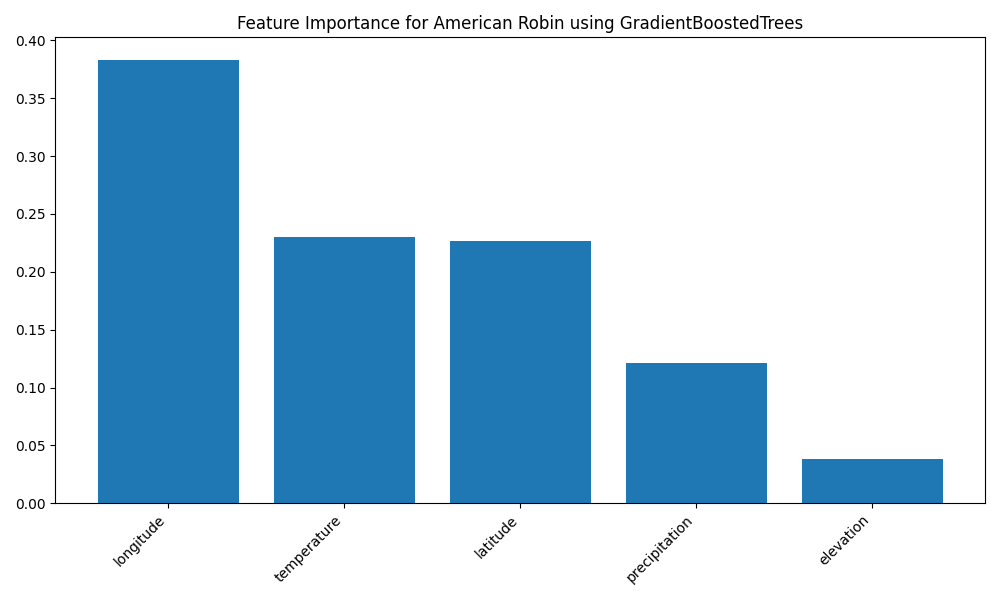}
\caption{Feature importance graph for the Gradient Boosted Trees model.}
\label{fig:robin_gbt_feat_imp}
\end{figure}

Since GBTs perform slightly worse than RFs, this discrepancy shows that our environmental features may include noise or unwanted information that may be skewing our results. RFs average this noise, hence provides better results than a GBT, a model which is more sensitive to noisy data. We will get back to this observation when we discuss future work in the conclusion section of our paper. Additionally, the features may separate presence/absence well, since RFs are effective when decision boundaries are clear. Hence, GBT may be overfitting based on some features, whereas RFs generate a more even spread of feature influence. Indeed, when we compare the importance of the features (see Figure~\ref{fig:robin_rf_feat_imp}), we notice that the RF implementation has a more even spread out of feature importance, a remark that shows that RFs may be building more generalized trees. However, when we compare the feature importance of RFs to Figure~\ref{fig:robin_gbt_feat_imp}, the GBTs may be overfitting based on longitude.

\begin{figure}[H]
\centering
\includegraphics[width=1\linewidth]{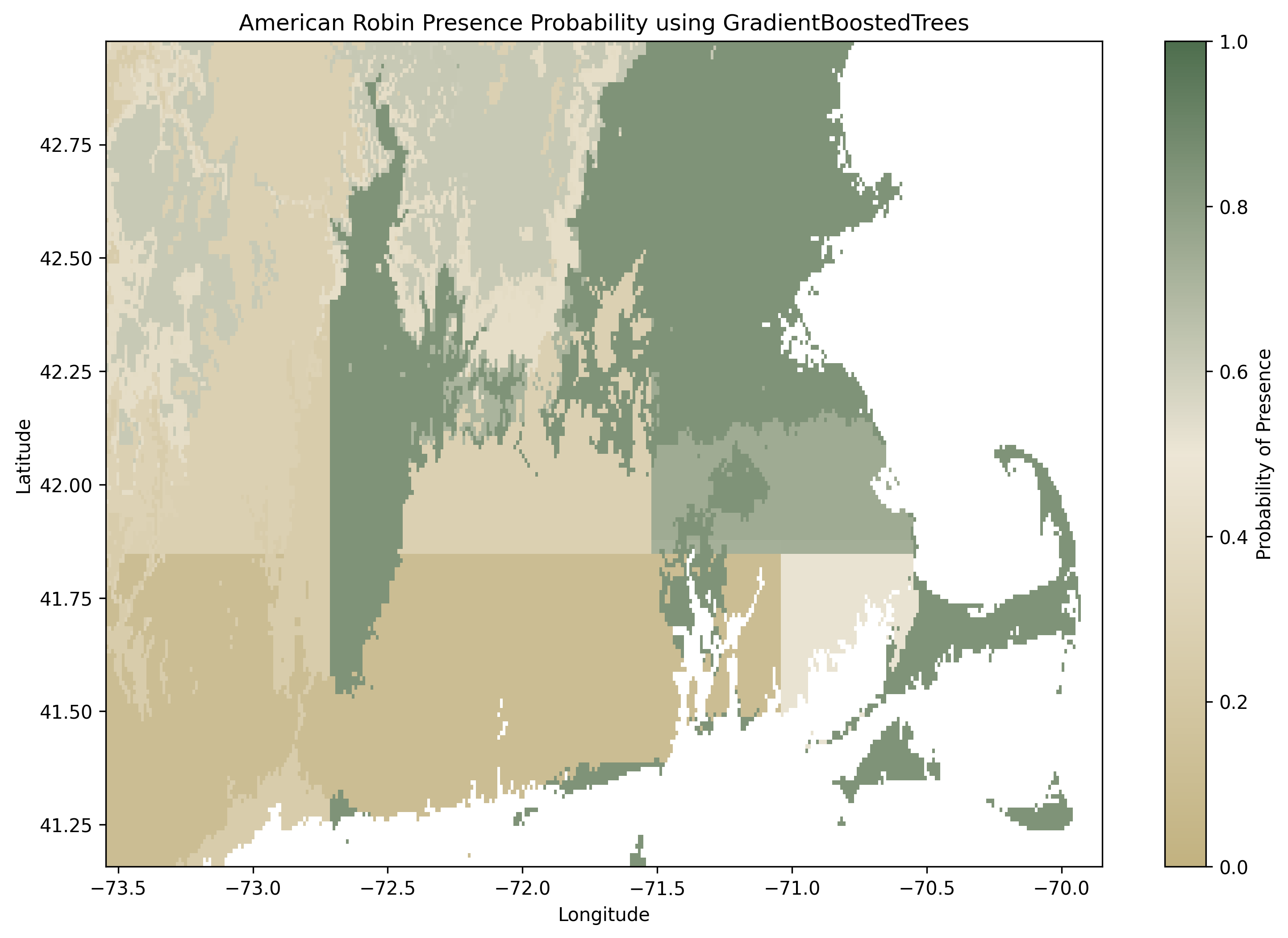}
\caption{Predicted distribution map for the American Robin using the Gradient Boosted Trees model.}
\label{fig:robin_gbt_map}
\end{figure}

\begin{figure}[H]
\centering
\includegraphics[width=1\linewidth]{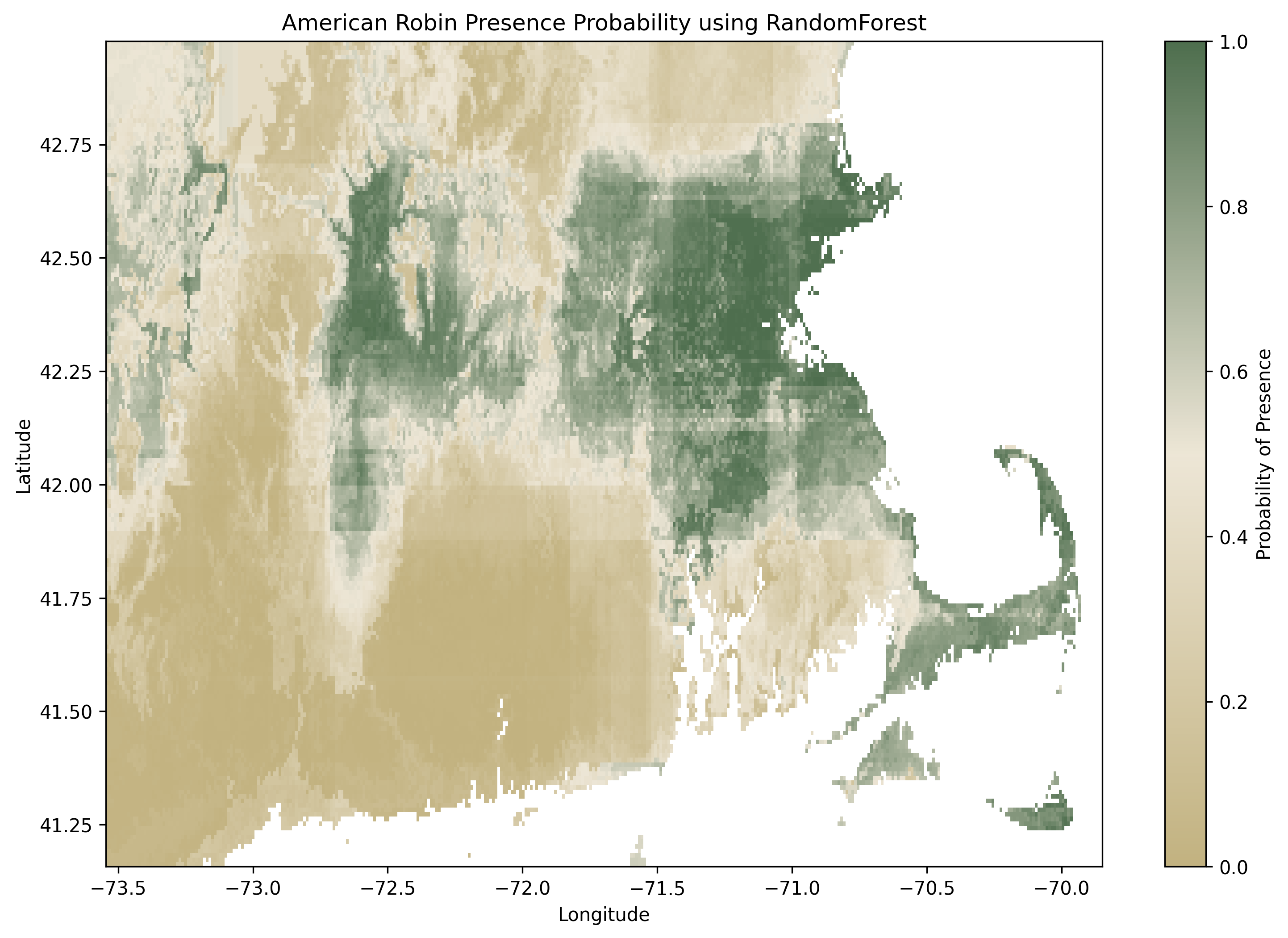}
\caption{Predicted distribution map for the American Robin using the Random Forest model.}
\label{fig:robin_rf_map}
\end{figure}

We note, however, that longitude seems to be the most important feature in determining where bird species are located. This could indicate that birds are looking to migrate towards warmer areas or coastal bodies. Take both Figure~\ref{fig:robin_rf_map} and Figure~\ref{fig:robin_gbt_map} as an example: notice how birds are strongly tied to coastal bodies as they provide food sources. With global warming, many food sources for birds may alter their presence and would thus affect bird migration.

\begin{figure}[H]
\centering
\includegraphics[width=1\linewidth]{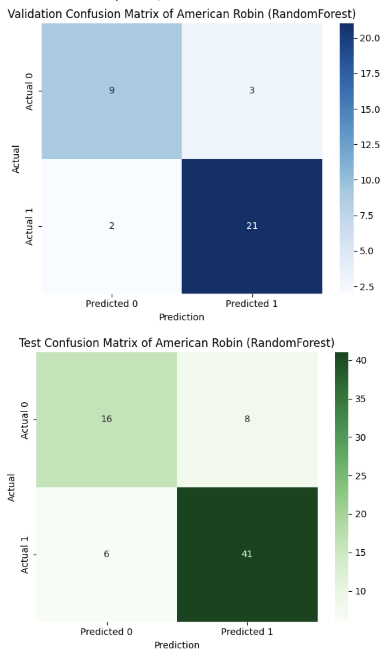}
\caption{Random Forest Confusion Matrix.}
\label{fig:rf_conf_mat}
\end{figure}

\begin{figure}[H]
\centering
\includegraphics[width=1\linewidth]{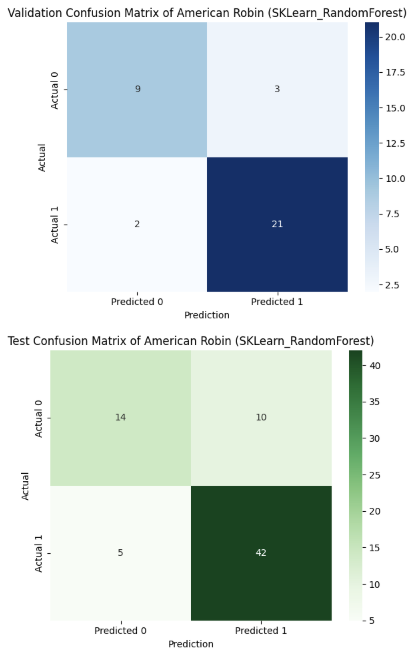}
\caption{Scikit-learn Random Forest Confusion Matrix.}
\label{fig:sk_rf_conf_mat}
\end{figure}

When compared to the baseline Scikit-learn models, we notice that our custom implementations are similar in accuracy. This insight shows that our models are classifying the data accurately. Comparing the confusion matrices between the two RF models as shown in Figure~\ref{fig:rf_conf_mat} and Figure~\ref{fig:sk_rf_conf_mat} we notice similarities in how well the model is predicting true positives and decently well in true negatives. This is most likely due to presence locations actually stemming from observations, whereas our pseudo-absent dataset is being generated and might not accurately reflect where birds may be absent.

\subsection{Convolutional Neural Network}   

Significant differences were found between the experimental results of the ResNet transfer learning approach to building the model and the custom CNN built from scratch. We will compare the two models in the following categories: performance comparison, classification patterns in the form of confusion matrices, feature importance, and discriminative power. Finally, we will be discussing the results of the two wholistically and their relevance to solving the problem of habitat shift modeling. 

\begin{figure}[H]
\centering
\includegraphics[width=1\linewidth]{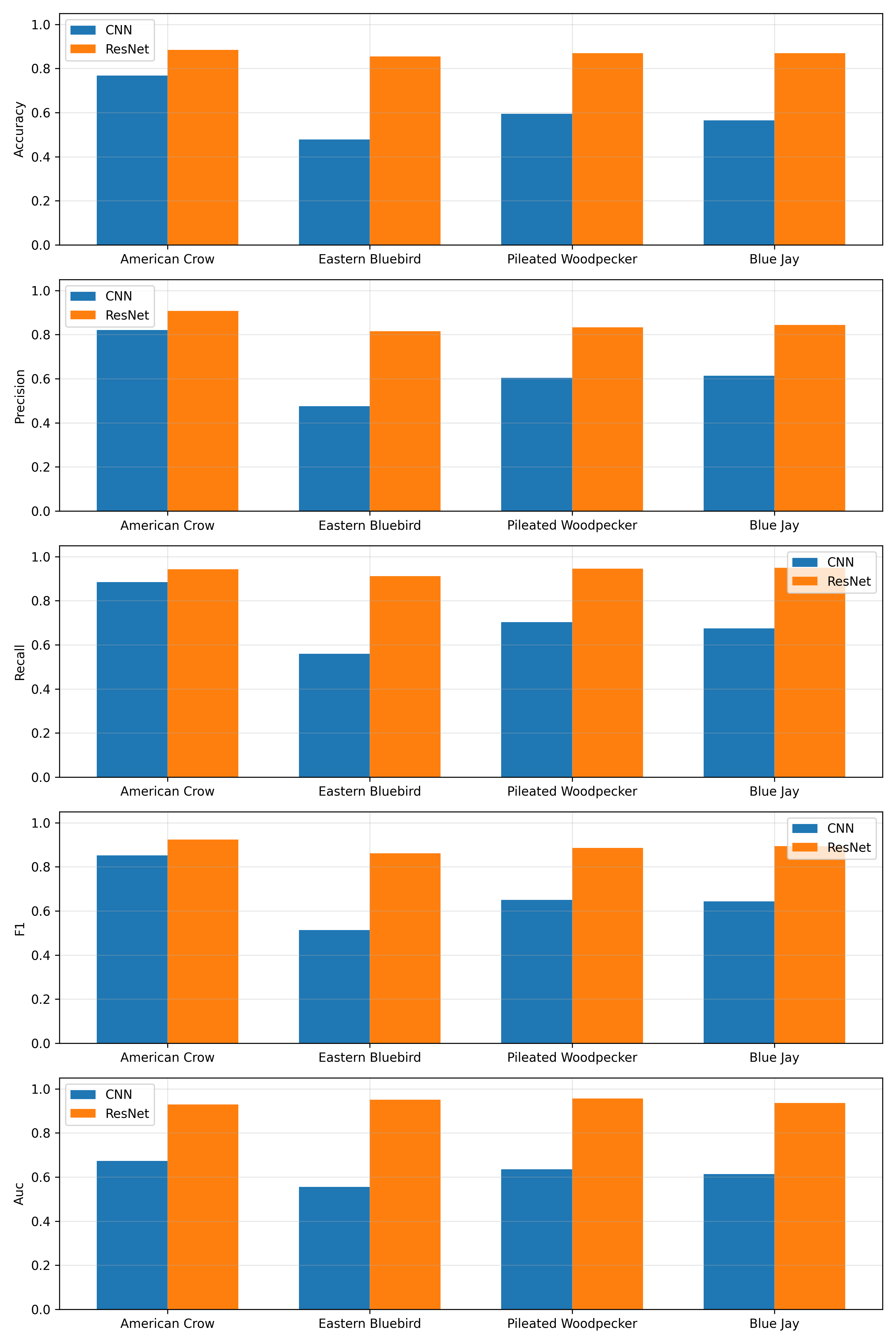}
\caption{ResNet vs CNN model results comparison}
\label{fig:model_compare}
\end{figure}

First, from Figure~\ref{fig:model_compare} we can see that the ResNet model outperforms the CNN consistently across all performance metrics. For precision, ResNet shown an average precision of 84.3\% versus 61.3\% for the custom CNN. For Recall, ResNet shows an average of 92.2\% compared to 69.3\% of the CNN. For the F1 score, ResNet averages a score of 88\% compared to the custom CNN of 65.8\%. Lastly the accuracy of ResNet shows an average of 91\% compared to the 61\% of the CNN. This substantial difference in all categories showcases the limitations of the CNN architecture to fully capture the complex landscape features that would influence bird distribution patterns during climate-induced range shifts. It also shows the fundamental advantages of using pre-trained ResNet on images as a backbone for a transfer learning on satellite images rather than a custom CNN without the image training of ResNet34. 

\begin{figure}[H]
\centering
\includegraphics[width=1\linewidth]{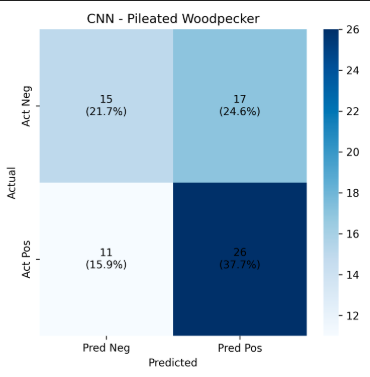}
\caption{Cnn Confusion Matrix for Pileated WoodPecker}
\label{fig:cnn_pile}
\end{figure}

\begin{figure}[H]
\centering
\includegraphics[width=1\linewidth]{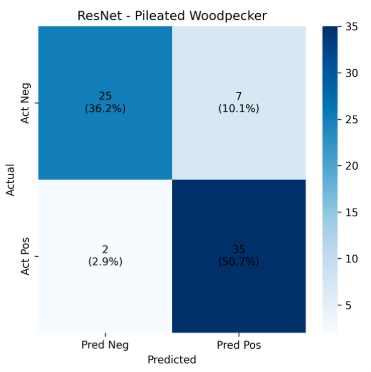}
\caption{ResNet Confusion Matrix for Pileated Woodpecker}
\label{fig:res_pile}
\end{figure}

Next, when comparing the confusion matrices generated from both models as shown from Figure~\ref{fig:cnn_pile} and Figure~\ref{fig:res_pile} about the pileated woodpecker. We can see that the ovverall accuracy differs greatly, 59.4\% for the custom CNN and 86.9\% for the ResNet. This might be the result of the insufficient model depth of the custom CNN's four block architecture compared to the ResNet architecture. It lacks the depth needed to capture complex ecological patterns. It also was not trained on anything other than the satellite image, which might be insufficient compared with ResNet which has already learned general image features. This allows the ResNet to adapt to the satellite images with relatively less additional data. 

\begin{figure}[H]
\centering
\includegraphics[width=1\linewidth]{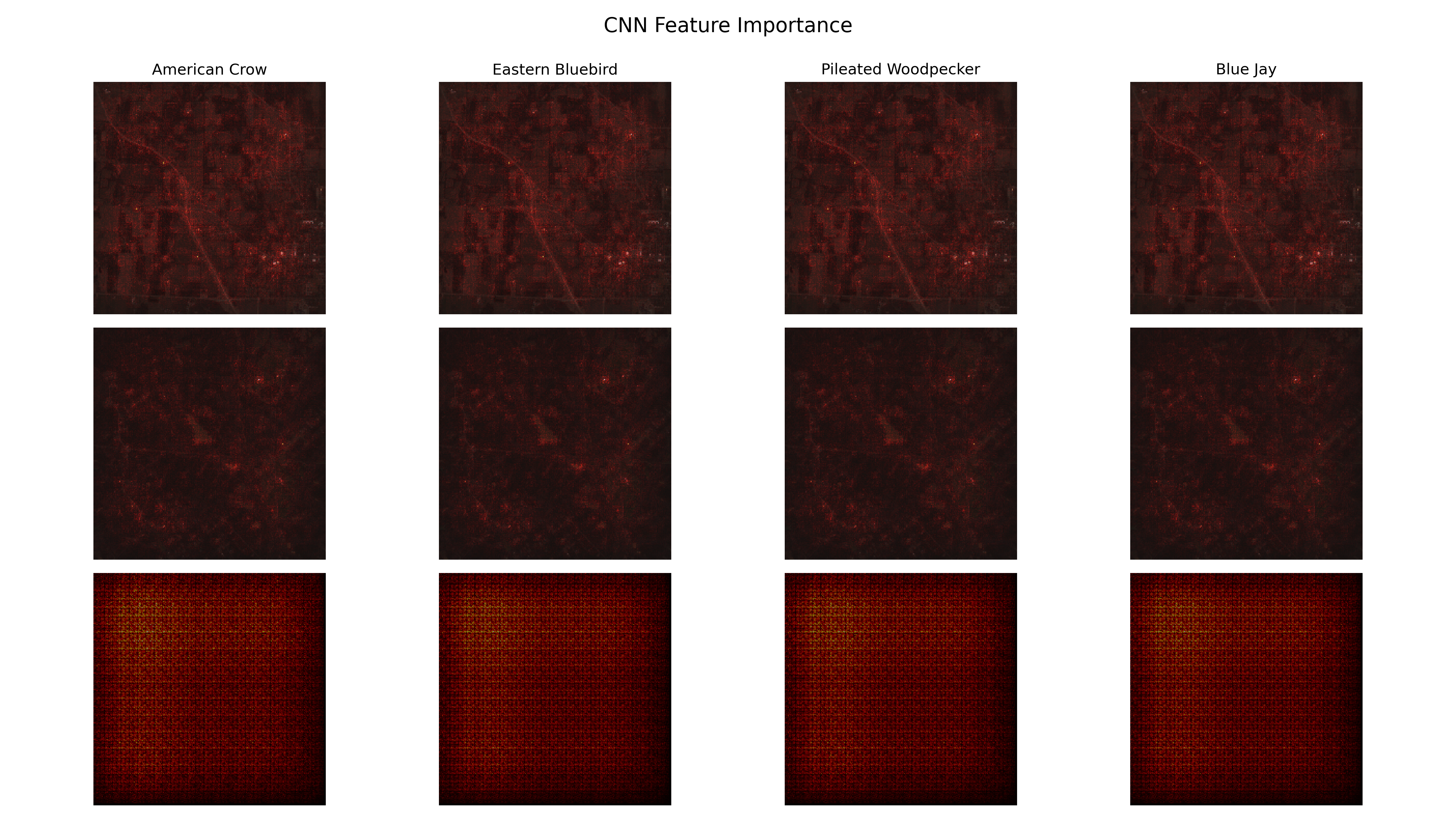}
\caption{CNN Feature importance}
\label{fig:cnn_feat}
\end{figure}

\begin{figure}[H]
\centering
\includegraphics[width=1\linewidth]{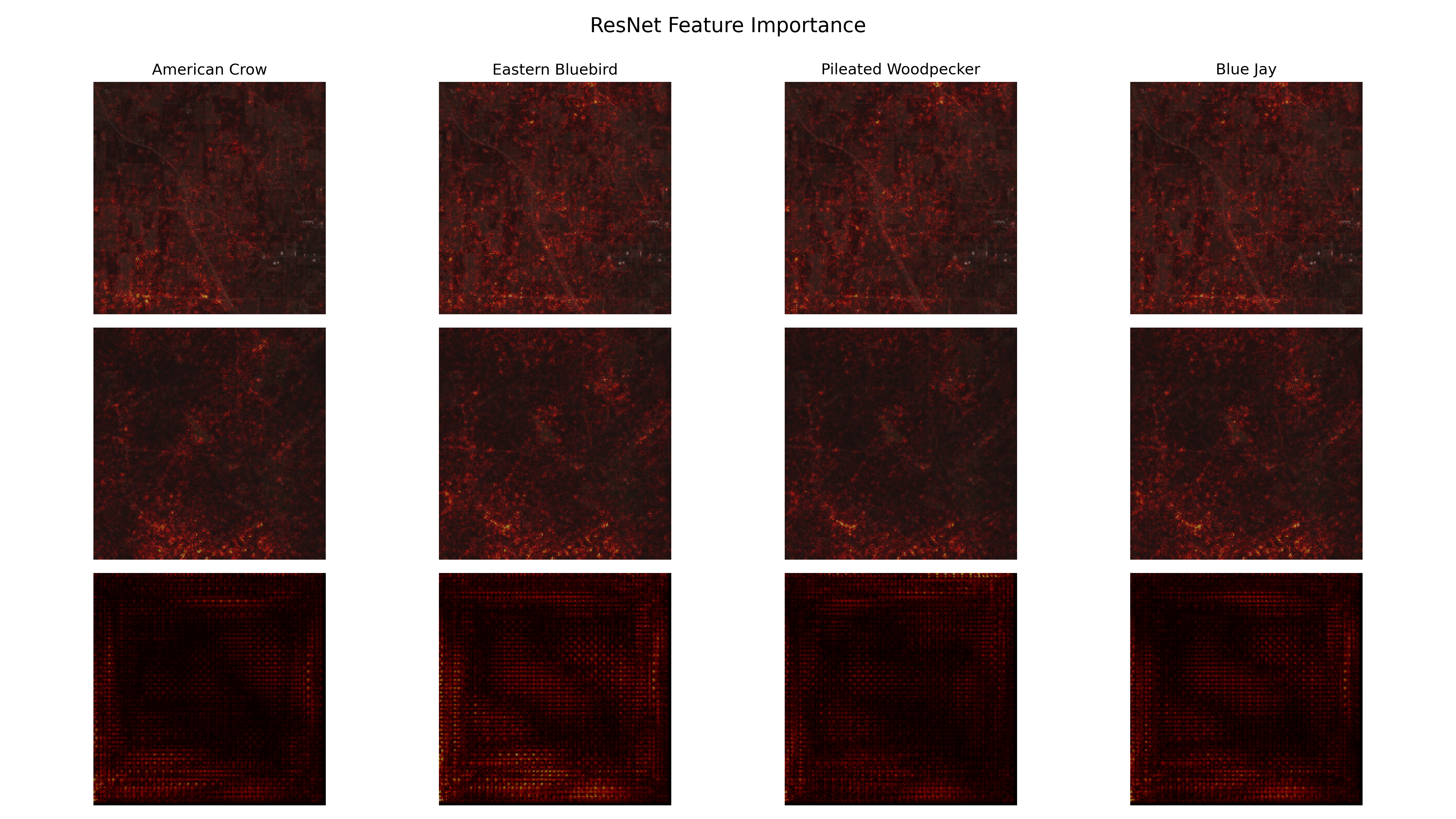}
\caption{ResNet Feature Importance}
\label{fig:res_feat}
\end{figure}

Thirdly, we must compare the feature importance between both the models itself. Looking at Figure~\ref{fig:cnn_feat} and Figure~\ref{fig:res_feat} we see that the custom CNN show more diffuse and less defined activation patterns compared to the ResNet one with a more structured and focused activations. ResNet was able to also demonstrate better spatial coherence where it was able to identify forest edges, water bodies, and vegetation patterns. These were features that the custom CNN was struggling to identify. The custom CNN had trouble with edge detection, where the transition zones should have shown as brighter activation patterns. This capability is relevant for predicting range shifts, as bird species follow habitat edges during migration. 

\begin{figure}[H]
\centering
\includegraphics[width=1\linewidth]{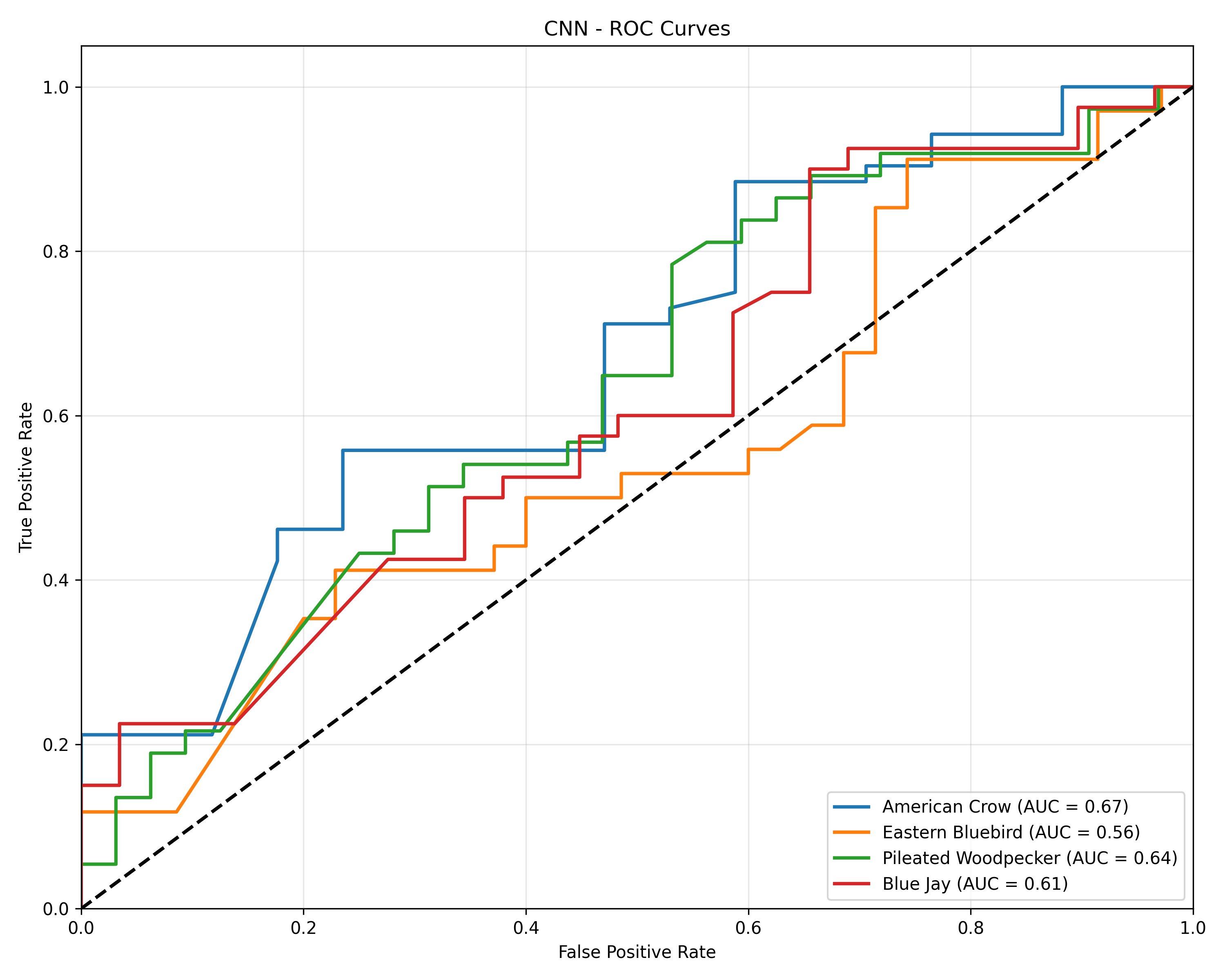}
\caption{CNN ROC curve}
\label{fig:cnn_roc}
\end{figure}

\begin{figure}[H]
\centering
\includegraphics[width=1\linewidth]{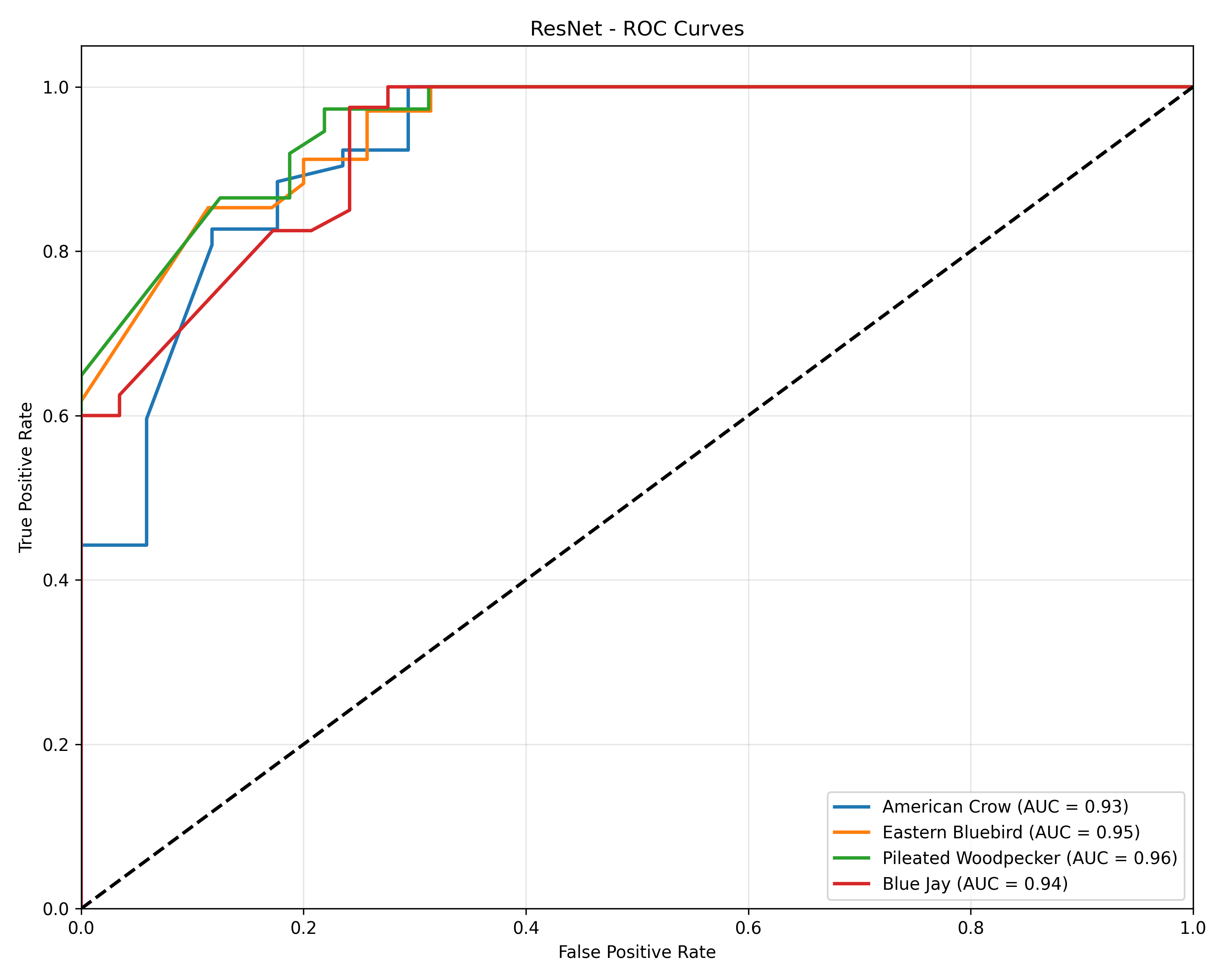}
\caption{ResNet ROC curve}
\label{fig:res_roc}
\end{figure}

Finally, we must discuss the difference in discriminative power between the custom CNN and ResNet. As shown in Figure~\ref{fig:cnn_roc} and Figure~\ref{fig:res_roc}, the custom CNN's ROC curves show greater irregularity and proximity to the random classifier line, which is shown as a straight diagonal dotted line. On the other hand, the ResNet's curve rises sharply to the top left corner. This is an indicator of excellent separation between positive and negative cases. Within the custom CNN, it shows highly variable performance across species, while ResNet maintains consistently high AUC scores for all bird species predicted. This would suggest that the ResNet has better generalization over diverse habitats preferences of the different bird species. 

\subsection{Comparing Tabular Data and ResNet/CNN}
The first apparent difference between the two approaches other than the method in which a prediction is made, is the feature interpretation between these approaches. Although tabular models identified longitude as the most significant characteristic indicating latitudinal migration patterns with a strong preference for the coast, CNN models used activation maps that emphasized landscape elements such as water bodies, vegetation patterns, and edge boundaries to capture visual patterns. However, it was true that birds were often found also residing near large bodies as supported by the results of both approaches. 

Interesting differences were also revealed by the classification patterns between the approaches. Tabular models demonstrated strong true positive identification across species with reasonable true negative performance, though potentially challenged by the artificial nature of the pseudo-absence data.

The ecological significance of both methods demonstrates how well they work together to model habitat shifts. Important climate and geographic factors influencing large-scale migration patterns were identified by tabular models that were in line with conventional climate envelope methodologies. By capturing landscape features that are difficult to measure in tabular data—particularly ecological transition zones that are crucial for migration—the CNN/ResNet models introduced a new dimension. Given that it captures both macro-level climate factors and micro-level landscape characteristics that impact habitat selection during climate-induced range shifts, an integrated approach that combines both methodologies may offer the most complete solution for predicting bird species distribution under climate change scenarios.

\section{Conclusion}

Our current work seems to be performing well, however, we note several areas for improvement. For one, we could consider other models like Logistic Regression, Support Vector Machine (SVM), or K-Nearest Neighbors \cite{Cutler2007}. Furthermore, we could look into changing the activation functions from a sigmoid to something like ReLU or tanh. We also want to look into combining our models to see whether the hybrid model performs better than having two separate models that predict based on features in image landscape and tabular data.

We also noticed that the GBT may be overfitting based on noise features in the dataset. We could improve our dataset for filtering for noise by providing more spatial resolutions of data points. Additionally, we could improve upon generating bird absence data points by gathering actual observation data for absent bird species' locations.

The more obvious limitation to our CNN/ResNet model is its small ecological dataset that the model was trained on\cite{Aodha2019}. Since ResNet was already pre-trained in general images, it performed better when confronted with new satellite images in comparison with CNN which only had the satellite images that it was provided. To improve this, more images might need to be used to train the custom CNN than the ResNet. 

Another limitation to the CNN/ResNet models is its reliance on RGB bands from Sentinel-2 imagery, constraining the model to patterns visible to the human eye rather than the full spectral information available. By incorporating non-visible spectral bands from Sentinel-2 imagery, such as near-infrared, short-wave infrared, and red-edge bands would enable the detection of vegetation health, moisture content, and biomass\cite{Kattenborn2019}. As a result, these variables that could effect the bird habitat would be included. 

Ultimately, we proposed a solution to predict bird species distributions. We hope that our approach will be used by ecologists in forecasting how bird habitats may shift in response to changing climate conditions. We wish our methods would contribute to policies to combat climate change and are inspired to further develop our approach to generate stronger models to predict where species are likely to relocate in the future.

\section{Contributions}

Min-Hong Shih was responsible for generating the CNN model; Emir Durakovic integrated the Tabular Data models. Both contributors contributed equally in writing the paper as well as helping to debug issues in the other's code.

\section{Code}

Our code can be found using this GitHub link: \url{https://github.com/Matt940624/Bird-Species-Distribution-Modeling}

\section{Appendix}

\end{document}